# Socially Assistive Robots as Decision Makers in the Wild: Insights from a Participatory Design Workshop

Ahmed, Eshtiak

Gamification Group, Tampere University, eshtiak.ahmed@tuni.fi

Cosio, Laura

Gamification Group, Tampere University, laura.cosio@tuni.fi

Hamari, Juho

Gamification Group, Tampere University, juho.hamari@tuni.fi

Buruk, Oğuz 'Oz'

Gamification Group, Tampere University, oguz.buruk@tuni.fi

Socially Assistive Robots (SARs) are becoming very popular every day because of their effectiveness in handling social situations. However, social robots are perceived as intelligent, and thus their decision-making process might have a significant effect on how they are perceived and how effective they are. In this paper, we present the findings from a participatory design study consisting of 5 design workshops with 30 participants, focusing on several decision-making scenarios of SARs in the wild. Through the findings of the PD study, we have discussed 5 directions that could aid the design of decision-making systems of SARs in the wild.

## 1 Introduction

We are living in an era where autonomous systems, such as robots, are a major part of our daily lives. These systems are developed in an attempt to make our day-to-day activities easier and less mundane. While being assistive in different tasks, robots have also been designed with capabilities of performing social interactions, hence they are categorized as Socially Assistive Robots (SAR) [1]. As these robots are deployed in scenarios where they assist humans in real-life situations, they need to possess a certain degree of autonomy and decision-making capabilities. The safety and efficiency of the interaction between humans and SARs depend heavily on these capabilities.

We conducted a participatory design (PD) workshop series to understand different aspects of designing a social robot that would accompany humans on nature visits. While the focus was on creating robot concepts including its appearance, behavior, and interactions, many interesting discussions came up on the decision-making of these robots. Nature or the wild as an environment for human-robot interaction (HRI) is vastly unpredictable and out of human control at times because of its variety of components acting in their own way. As a result, a SAR will need to process a lot more unknown information in the wild compared to a controlled environment. In this paper, we will briefly discuss the method and findings of the PD study with a keen focus on how the findings demonstrate decision-making scenarios for SARs in the wild.

In the workshop, findings from this PD study will potentially add a new dimension of discussion to answer questions like:
- Which decisions in which scenarios should be made by SAR?
- How to build trust in assistive scenarios with SARs?
- How to handle unpredictable situations created by SAR's limitations?
- How can SARs adapt to different users in different scenarios?

On the other hand, discussions around SAR embodiment, communication transparency, moral reasoning, and imperfections will help us better understand the design space of a SAR companion in the wild. This in turn will provide us with a different perspective for strengthening our research on the human-robot-nature relationship.

## 2 METHOD

We organized a total of five (3 Atom + 1 Synthesis + 1 Fusion) workshops involving 30 participants who represented different stakeholder groups of the design. Atom workshops focus on the different aspects of the robotic companion for nature exploration, in our case the appearance and behavior of the companion as well as the environment of the interaction. In each Atom workshop, participants used the brainstorming method to come up with the 3 most valued themes (MVTs) for the design of one specific aspect, resulting in a total of 9 MVTs. In the Synthesis workshop, the participants were divided into 5 groups and brainstormed to create in-depth concepts merging the learnings about the different facets of robotic companion design and presenting them to get feedback on improving them. The concepts created in the Synthesis workshop were taken to the Fusion workshop where the participants improved them based on feedback. In addition to that, low-fidelity prototyping was done by creating storyboards [2] and video sketching [3].

## 3 CONCEPTS

5 concepts were designed by attempting to incorporate the MVTs. 3 of those concepts are included here as they have focused on several decision-making aspects of SARs. These 3 concepts are visualized in Figure 1. Unique aspects of the concepts are discussed in the following subsections.

### 3.1 Zeus (Fig 1 - A1, A2)

The most notable design choices for this concept were the safety behavior (e.g., the robot scanned a frozen lake and suggested the user not walk on it) of the robot and the user's lack of trust in the robot's suggestions (e.g., the user did not trust the robot when it advised them to not walk on the frozen lake and walked on it anyway). It also showed that the robot did not leave the user alone even if the user would not listen to suggestions. If the user chose to ignore suggestions, the robot would respect this and not continue to bring it up. Through ignorance, the user fell into a frozen lake and started drowning. The robot reacted to this situation by throwing a buoy at the user and then dragging the user ashore.

### 3.2 G3 (Fig 1 - C3)

The most important design choice of G3 is its ability to change shapes between a backpackable object and a full-fledged mobile robot. This makes the robot easy to use because if anything happens to the robot during outdoor activity, it can be easily carried back home. The robot can assess the edibility of fruits or any other thing in nature and helps the user find safe food. This concept also shows that robots can make mistakes, such as waking the user up in the middle of the night as it sensed danger. While it is important for the robot to have protective behavior, its judgment cannot be trusted in all situations as it might make the interaction unreliable.

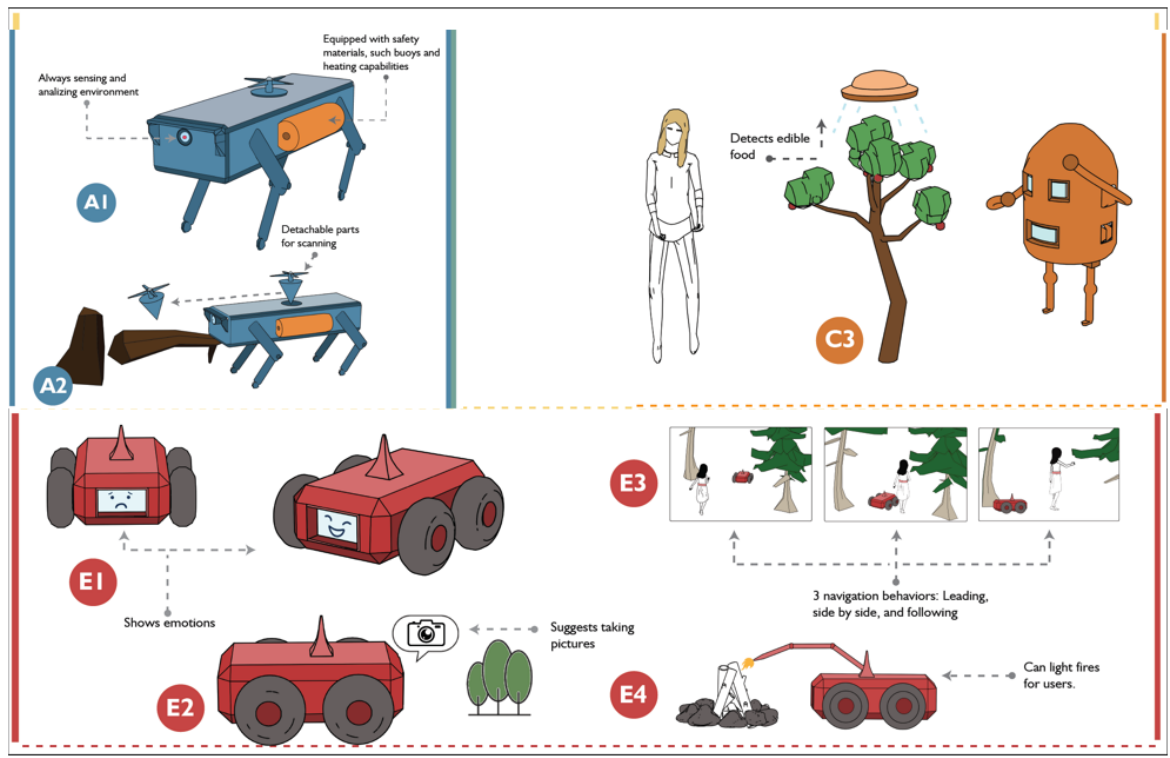

Figure 1: Fusion workshop concepts (Zeus (A), G3 (C), and Boombot (E)).

### 3.3 Boombot (Fig 1 - E1, E2, E3, E4)

This concept shows that the robot exhibits different types of emotions in different scenarios which can lead to a deeper connection with the user and better overall interaction. It has three walking modes, showing the path to the user, following the user, and walking side by side with the user. The robot can recognize the user's preference for social activities and suggests taking pictures for social media. The robot also uses common sense to not try to go immediately home when it starts raining, so it suggested going to a shelter. It is also capable of safely lighting a fire to create a warmer environment for the user.

## 4  DECISION-MAKING IN THE WILD THROUGH ROBOTS

The concepts created in the Fusion workshop demonstrated several scenarios in the wild where the robotic companion had to make conscious decisions and take actions accordingly, especially to improve the interaction. Through careful analysis of the concepts, the following directions appear to be beneficial for designing decision systems for autonomous SARs that are intended to operate in the wild.

### 4.1 Adapting to human behavior and social norms

A SAR should be able to imitate human behavior and social norms to become more relatable to humans. When people interact with SARs, they will subconsciously expect robots to understand human behavior and act accordingly [4]. It has been reported that social and moral norms of humans are expected from robots in social situations and such robots must be able to understand, embody, act upon, and apply social and moral rules [5]. In concept Zeus, the robot acted as a companion to the user and respected their opinion even though they were not taking suggestions. In the concept Boombot, the robot understood the user's social status and suggested taking pictures for social media.

### 4.2 Mishap management

In nature, unwanted situations might occur and the robot should be able to react to such situations. At times, the wellness and survival of the user might depend on how the robot reacts to a situation [6]. In concept Zeus, the robot reacted rapidly when the user's life was in danger and as a result, it was able to save the user. In the concept Boombot, the robot reacted in time to take the user to a shelter before it started raining heavily.

### 4.3 Ensuring safety

SARs should always be aware of the interaction environment and ensure the safety of the interaction, especially in nature. They should have the ability to scan and survey the area to know about potential threats so that they can act accordingly. SARs need to be concerned about the safety of the user and exhibit appropriate protective behavior through intelligent decision-making. In concept G3, the robot first analyzes if a fruit in the wild is edible before it lets the user eat them.

### 4.4 Embracing the unpredictability of nature

Without being careful, dangerous situations might occur in the wild and as a companion in nature, the SAR might be able to play a very important role in such situations. So, it is important for the robot to have situational awareness and reactions [7]. In nature, many things are out of control of the humans, for example, in concept G3, the robot suddenly sensed that it is going to rain. In this situation, the robot needed to have the ability to decide what to do.

### 4.5 Building trust in the SAR's ability

The level of expectation from a SAR is crucial in building trust and comfort around it. Although recent technology and AI have made robots much more capable in terms of social encounters, recognizing emotions, and being able to evoke emotions in humans, there are limitations to their capabilities [8]. In Zeus (group 1), the robot failed to understand the user and conveyed important information in a way that did not catch the attention of the user which led to an unwanted situation (drowning in the frozen lake). SARs as well as their human counterparts need to be aware of their limitations and act to their strengths to make decisions, especially in uncontrollable environments such as the wild.

## 5 Conclusion

Autonomous SARs are becoming an integral part of human lives and as a result, their decision-making abilities have become very important. Through the findings from a PD study, this paper reflected on 5 directions of understanding the decision-making of SARs in the wild. These directions might aid in understanding several questions in the field, the extent of decision-making of SARs, building trust in assistive scenarios, imperfections of SARs as decision-makers, and their adaptation to specific populations.